\documentclass[twoside,11pt]{article} 
\usepackage{jmlr2e} 

\jmlrheading{}{}{}{}{}{Marc Claesen, Jaak Simm, Dusan Popovic, Yves Moreau and Bart De Moor}

\ShortHeadings{Optunity}{Claesen, Simm, Popovic, Moreau and De Moor}
\firstpageno{1}

\usepackage{amsmath}
\usepackage{tikz}
\usepackage{ifpdf}
\usepackage{multirow}
\usepackage{afterpage}
\usepackage{booktabs}
\usepackage{fancyvrb}
\usepackage{color}
\usepackage[colorlinks=false,allbordercolors={1 1 1}]{hyperref}
\usepackage{listings}

\newcommand{\train}{\mathbf{X}^{(tr)}}
\newcommand{\test}{\mathbf{X}^{(te)}}
\newcommand{\obj}{\mathcal{F}}
\DeclareMathOperator*{\argmin}{arg\,min}

\definecolor{mygreen}{rgb}{0,0.6,0}
\definecolor{mygray}{rgb}{0.5,0.5,0.5}
\definecolor{bggray}{rgb}{0.95,0.95,0.95}
\definecolor{mymauve}{rgb}{0.58,0,0.82}
\lstdefinestyle{Py}{
    language={Python}, 
    moredelim=**[is][\slshape\bfseries]{`}{`},
    moredelim=**[is][\color{mygreen}]{!}{!},
}
\begin{document}

\title{Easy Hyperparameter Search Using Optunity}

\author{\name Marc Claesen \email marc.claesen@esat.kuleuven.be  \\ 
\name Jaak Simm \email jaak.simm@esat.kuleuven.be   \\ 
\name Dusan Popovic \email dusan.popovic@esat.kuleuven.be   \\ 
\name Yves Moreau \email yves.moreau@esat.kuleuven.be \\
\name Bart De Moor \email bart.demoor@esat.kuleuven.be  \\ 
\addr KU Leuven, Department of Electrical Engineering (ESAT) \\
STADIUS Center for Dynamical Systems, Signal Processing and Data Analytics \\ 
iMinds, Department of Medical Information Technologies \\
Kasteelpark Arenberg 10, box 2446 \\
3001 Leuven, Belgium
}


\maketitle

\begin{abstract}
    Optunity is a free software package dedicated to hyperparameter optimization. It contains various types of solvers, ranging from undirected methods to direct search, particle swarm and evolutionary optimization. The design focuses on ease of use, flexibility, code clarity and interoperability with existing software in all machine learning environments. Optunity is written in Python and contains interfaces to environments such as R and MATLAB.  Optunity uses a BSD license and is freely available online at \texttt{\url{http://www.optunity.net}}.
\end{abstract}


\section{Introduction}
Many machine learning tasks aim to train a model $\mathcal{M}$ which minimizes some loss function $\mathcal{L}(\mathcal{M}\ |\ \test)$ on given test data $\test$. A model is obtained via a learning algorithm $\mathcal{A}$ which uses a training set $\train$ and solves some optimization problem. The learning algorithm $\mathcal{A}$ may itself be parameterized by a set of hyperparameters $\lambda$, e.g. $\mathcal{M} = \mathcal{A}(\train\ |\ \lambda)$.  Hyperparameter search -- also known as tuning -- aims to find a set of hyperparameters $\lambda^*$, such that the learning algorithm yields an optimal model $\mathcal{M}^*$ that minimizes $\mathcal{L}(\mathcal{M}\ |\ \test)$:
\begin{equation}
\lambda^* = \argmin_{\lambda} \mathcal{L}\big(\mathcal{A}(\train\ |\ \lambda)\ |\ \test\big) = \argmin_{\lambda} \obj(\lambda\ |\ \mathcal{A},\ \train, \test,\ \mathcal{L}) \label{equation}
\end{equation}
In the context of tuning, $\obj$ is the objective function and $\lambda$ is a tuple of hyperparameters (optimization variables). The learning algorithm $\mathcal{A}$ and data sets $\train$ and $\test$ are known. Depending on the learning task, $\train$ and $\test$ may be labeled and/or equal to each other. The objective function often has a constrained domain (for example regularization terms must be positive) and is assumed to be expensive to evaluate, black-box and non-smooth.

Tuning hyperparameters is a recurrent task in many machine learning approaches. Some common hyperparameters that must be tuned are related to kernels, regularization, learning rates and network architecture. Tuning can be necessary in both supervised and unsupervised settings and may significantly impact the resulting model's performance. 

General machine learning packages typically provide only basic tuning methods like grid search. The most common tuning approaches are grid search and manual tuning \citep{hsu2003practical, hinton2012practical}. Grid search suffers from the curse of dimensionality when the number of hyperparameters grows large while manual tuning requires considerable expertise which leads to poor reproducibility, particularly when many hyperparameters are involved. 

\section{Optunity}
Our software is a Swiss army knife for hyperparameter search. Optunity offers a series of configurable optimization methods and utility functions that enable efficient hyperparameter optimization. Only a handful of lines of code are necessary to perform tuning. Optunity should be used in tandem with existing machine learning packages that implement learning algorithms. The package uses a BSD license and is simple to deploy in any environment. Optunity has been tested in Python, R and MATLAB on Linux, OSX and Windows.

\subsection{Functional overview}
Optunity provides both simple routines for lay users and expert routines that enable fine-grained control of various aspects of the solving process. Basic tuning can be performed with minimal configuration, requiring only an objective function, an upper limit on the number of evaluations and box constraints on the hyperparameters to be optimized.

The objective function must be defined by the user. It takes a hyperparameter tuple $\lambda$ and typically involves three steps: (i) training a model $\mathcal{M}$ with $\lambda$, (ii) use $\mathcal{M}$ to predict a test set (iii) compute some score or loss based on the predictions. In unsupervised tasks, the separation between (i) and (ii) need not exist, for example in clustering a data set.

Tuning involves a series of function evaluations until convergence or until a predefined maximum number of evaluations is reached. Optunity is capable of vectorizing evaluations in the working environment to speed up the process at the end user's volition.

Optunity additionally provides $k$-fold cross-validation to estimate the generalization performance of supervised modeling approaches. The cross-validation implementation can account for strata and clusters.\footnote{Instances in a stratum should be spread across folds. Clustered instances must remain in a single fold.} Finally, a variety of common quality metrics is available.

The code example below illustrates tuning an SVM with scikit-learn and Optunity.\footnote{We assume the correct imports are made and \texttt{data} and \texttt{labels} contain appropriate content.}

\begin{lstlisting}[style=Py, frame=none, xleftmargin=3.5ex, escapeinside={(*@}{@*)}, belowskip=-2.2\medskipamount]
!@optunity.cross_validated!(x=(*@\textcolor{red}{data}@*), y=(*@\textcolor{red}{labels}@*), num_folds=10, num_iter=2)
def (*@\textcolor{blue}{svm\_auc}@*)(x_train, y_train, x_test, y_test, (*@\textcolor{blue}{C}@*), (*@\textcolor{blue}{gamma}@*)):
    model = sklearn.svm.SVC(C=(*@\textcolor{blue}{C}@*), gamma=(*@\textcolor{blue}{gamma}@*)).fit(x_train, y_train)
    decision_values = model.decision_function(x_test)
    return !optunity.metrics.roc_auc!(y_test, decision_values)

optimal_pars, _, _ = !optunity.maximize!((*@\textcolor{blue}{svm\_auc}@*), num_evals=100, (*@\textcolor{blue}{C=[0, 10]}@*), (*@\textcolor{blue}{gamma=[0, 1]}@*)) (*@\label{optunity-maximize}@*)
optimal_model = sklearn.svm.SVC(**optimal_pars).fit((*@\textcolor{red}{data}@*), (*@\textcolor{red}{labels}@*))
\end{lstlisting}

The objective function as per Equation~\eqref{equation} is defined on lines 1 to 5, where $\lambda = (C, \gamma)$, $\mathcal{A}$ is the SVM training algorithm and $\mathcal{L}$ is area under the ROC curve. We use $2\times$ iterated 10-fold cross-validation to estimate area under the ROC curve. Up to $100$ hyperparameter tuples are tested within the box constraints $0 < C < 10$ and $0 < \gamma < 1$ on line \ref{optunity-maximize}.

\subsection{Available solvers}
{\noindent}Optunity provides a wide variety of solvers, ranging from basic, undirected methods like grid search and random search \citep{bergstra2012random} to evolutionary methods such as particle swarm optimization \citep{kennedy2010particle} and the covariance matrix adaptation evolutionary strategy (CMA-ES) \citep{hansen2001completely}. Finally, we provide the Nelder-Mead simplex \citep{nelder1965simplex}, which is useful for local search after a good region has been determined. Optunity's current default solver is particle swarm optimization, as our experiments have shown it to perform well for a large variety of tuning tasks involving various learning algorithms. Additional solvers will be incorporated in the future.
 
\subsection{Software design and implementation}

The design philosophy of Optunity prioritizes code clarity over performance. This is justified by the fact that objective function evaluations constitute the real performance bottleneck. 

In contrast to typical Python packages, we avoid dependencies on big packages like NumPy/SciPy and scikit-learn to facilitate users working in non-Python environments (sometimes at the cost of performance). To prevent issues for users that are unfamiliar with Python, care is taken to ensure all code in Optunity works out of the box on any Python version above 2.7, without requiring tools like \texttt{2to3} to make explicit conversions. Optunity has a single dependency on DEAP \citep{fortin2012deap} for the CMA-ES solver. 

\begin{figure}[!h]
  \centering 
      \includegraphics[width=0.8\textwidth]{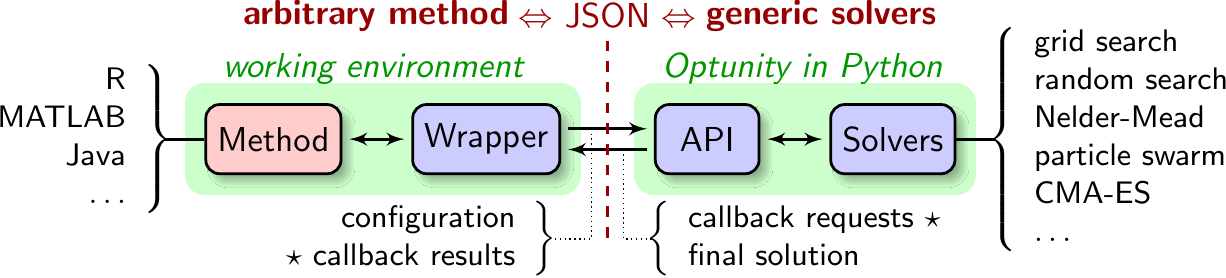} 
  \caption{Integrating Optunity in non-Python environments.}\label{fig:workflow}
\end{figure}

A key aspect of Optunity's design is interoperability with external environments. This requires bidirectional communication between Optunity's Python back-end ($\mathcal{O}$) and the external environment ($\mathcal{E}$) and roughly involves three steps: (i) $\mathcal{E}\rightarrow\mathcal{O}$ solver configuration, (ii) $\mathcal{O}\leftrightarrow\mathcal{E}$ objective function evaluations and (iii) $\mathcal{O}\rightarrow\mathcal{E}$ solution and solver summary. To this end, Optunity can do straightforward communication with any environment via sockets using JSON messages as shown in Figure~\ref{fig:workflow}. Only some information must be communicated, big objects like data sets are never exchanged. To port Optunity to a new environment, a thin wrapper must be implemented to handle communication.

\subsection{Documentation}
Code is documented using Sphinx and contains many doctests that can serve as both unit tests and examples of the associated functions. 
Our website contains API documentation, user documentation and a wide range of examples to illustrate all aspects of the software. 
The examples involve various packages, including scikit-learn \citep{pedregosa2011scikit}, OpenCV \citep{opencv_library} and Spark's MLlib \citep{zaharia2010spark}.

\newpage
\subsection{Collaborative and future development}
Collaborative development is organized via GitHub.\footnote{We maintain the following subdomains for convenience: \texttt{http://}$\{$\href{http://builds.optunity.net}{builds}, \href{http://docs.optunity.net}{docs}, \href{http://git.optunity.net}{git}, \href{http://issues.optunity.net}{issues}$\}$\texttt{.optunity.net}.} The project's master branch is kept stable and is subjected to continuous integration tests using Travis CI. 
We recommend prospective users to clone the master branch for the most up-to-date stable version of the software. Bug reports and feature requests can be filed via issues on GitHub.

Future development efforts will focus on wrappers for Java, Julia and C/C++. This will make Optunity readily available in all main environments related to machine learning. We additionally plan to incorporate Bayesian optimization strategies \citep{jones1998efficient}. 


\section{Related work}

A number of software solutions exist for hyperparameter search. HyperOpt offers random search and sequential model-based optimization \citep{bergstra2013hyperopt}. Some packages dedicated to Bayesian approaches include Spearmint \citep{snoek2012practical}, DiceKriging \citep{roustant2012dicekriging} and BayesOpt \citep{martinez2014bayesopt}. Finally, ParamILS is a command-line-only tuning framework providing iterated local search \citep{hutter2009paramils}. 

Optunity distinguishes itself from existing packages by exposing a variety of fundamentally different solvers. This matters because the no free lunch theorem suggests that no single approach is best in all settings \citep{wolpert1997no}. Additionally, Optunity is easy to integrate in various environments and features a very simple API.

\acks{This research was funded via the following channels: 
\begin{itemize}
\setlength\itemsep{0.2em}
\item Research Council KU Leuven: GOA/10/09 MaNet, CoE PFV/10/016 SymBioSys; 
\item Flemish Government: FWO: projects:  G.0871.12N (Neural circuits); IWT: TBM-Logic Insulin(100793), TBM Rectal Cancer(100783), TBM IETA(130256), O\&O ExaScience Life Pharma, ChemBioBridge, PhD grants (specifically 111065); Industrial Research fund (IOF): IOF/HB/13/027 Logic Insulin; iMinds Medical Information Technologies SBO 2014; VLK Stichting E. van der Schueren: rectal cancer
\item Federal Government: FOD: Cancer Plan 2012-2015 KPC-29-023 (prostate)
\item COST: Action: BM1104: Mass Spectrometry Imaging
\end{itemize}}

\bibliography{bibliography}

\begin{thebibliography}{17}
\providecommand{\natexlab}[1]{#1}
\providecommand{\url}[1]{\texttt{#1}}
\expandafter\ifx\csname urlstyle\endcsname\relax
  \providecommand{\doi}[1]{doi: #1}\else
  \providecommand{\doi}{doi: \begingroup \urlstyle{rm}\Url}\fi

\bibitem[Bergstra and Bengio(2012)]{bergstra2012random}
James Bergstra and Yoshua Bengio.
\newblock Random search for hyper-parameter optimization.
\newblock \emph{Journal of Machine Learning Research}, 13\penalty0
  (1):\penalty0 281--305, 2012.

\bibitem[Bergstra et~al.(2013)Bergstra, Yamins, and Cox]{bergstra2013hyperopt}
James Bergstra, Dan Yamins, and David~D Cox.
\newblock Hyperopt: A python library for optimizing the hyperparameters of
  machine learning algorithms.
\newblock In \emph{Proceedings of the 12th Python in Science Conference}, pages
  13--20. SciPy, 2013.

\bibitem[Bradski(2000)]{opencv_library}
G.~Bradski.
\newblock The {OpenCV} library.
\newblock \emph{Dr. Dobb's Journal of Software Tools}, 2000.
\newblock URL
  \url{http://www.drdobbs.com/open-source/the-opencv-library/184404319}.

\bibitem[Fortin et~al.(2012)Fortin, Rainville, Gardner, Parizeau, Gagn{\'e},
  et~al.]{fortin2012deap}
F{\'e}lix-Antoine Fortin, De~Rainville, Marc-Andr{\'e}~Gardner Gardner, Marc
  Parizeau, Christian Gagn{\'e}, et~al.
\newblock {DEAP}: Evolutionary algorithms made easy.
\newblock \emph{Journal of Machine Learning Research}, 13\penalty0
  (1):\penalty0 2171--2175, 2012.

\bibitem[Hansen and Ostermeier(2001)]{hansen2001completely}
Nikolaus Hansen and Andreas Ostermeier.
\newblock Completely derandomized self-adaptation in evolution strategies.
\newblock \emph{Evolutionary computation}, 9\penalty0 (2):\penalty0 159--195,
  2001.

\bibitem[Hinton(2012)]{hinton2012practical}
Geoffrey~E Hinton.
\newblock A practical guide to training restricted boltzmann machines.
\newblock In \emph{Neural Networks: Tricks of the Trade}, pages 599--619.
  Springer, 2012.

\bibitem[Hsu et~al.(2003)Hsu, Chang, Lin, et~al.]{hsu2003practical}
Chih-Wei Hsu, Chih-Chung Chang, Chih-Jen Lin, et~al.
\newblock A practical guide to support vector classification, 2003.

\bibitem[Hutter et~al.(2009)Hutter, Hoos, Leyton-Brown, and
  St{\"u}tzle]{hutter2009paramils}
Frank Hutter, Holger~H Hoos, Kevin Leyton-Brown, and Thomas St{\"u}tzle.
\newblock {ParamILS}: an automatic algorithm configuration framework.
\newblock \emph{Journal of Artificial Intelligence Research}, 36\penalty0
  (1):\penalty0 267--306, 2009.

\bibitem[Jones et~al.(1998)Jones, Schonlau, and Welch]{jones1998efficient}
Donald~R Jones, Matthias Schonlau, and William~J Welch.
\newblock Efficient global optimization of expensive black-box functions.
\newblock \emph{Journal of Global optimization}, 13\penalty0 (4):\penalty0
  455--492, 1998.

\bibitem[Kennedy(2010)]{kennedy2010particle}
James Kennedy.
\newblock Particle swarm optimization.
\newblock In \emph{Encyclopedia of Machine Learning}, pages 760--766. Springer,
  2010.

\bibitem[Martinez-Cantin(2014)]{martinez2014bayesopt}
Ruben Martinez-Cantin.
\newblock {BayesOpt}: A {Bayesian} optimization library for nonlinear
  optimization, experimental design and bandits.
\newblock \emph{arXiv preprint arXiv:1405.7430}, 2014.

\bibitem[Nelder and Mead(1965)]{nelder1965simplex}
John~A Nelder and Roger Mead.
\newblock A simplex method for function minimization.
\newblock \emph{The computer journal}, 7\penalty0 (4):\penalty0 308--313, 1965.

\bibitem[Pedregosa et~al.(2011)Pedregosa, Varoquaux, Gramfort, Michel, Thirion,
  Grisel, Blondel, Prettenhofer, Weiss, Dubourg, et~al.]{pedregosa2011scikit}
Fabian Pedregosa, Ga{\"e}l Varoquaux, Alexandre Gramfort, Vincent Michel,
  Bertrand Thirion, Olivier Grisel, Mathieu Blondel, Peter Prettenhofer, Ron
  Weiss, Vincent Dubourg, et~al.
\newblock Scikit-learn: Machine learning in {Python}.
\newblock \emph{Journal of Machine Learning Research}, 12:\penalty0 2825--2830,
  2011.

\bibitem[Roustant et~al.(2012)Roustant, Ginsbourger, Deville,
  et~al.]{roustant2012dicekriging}
Olivier Roustant, David Ginsbourger, Yves Deville, et~al.
\newblock {DiceKriging}, {DiceOptim}: Two {R} packages for the analysis of
  computer experiments by kriging-based metamodeling and optimization.
\newblock 2012.

\bibitem[Snoek et~al.(2012)Snoek, Larochelle, and Adams]{snoek2012practical}
Jasper Snoek, Hugo Larochelle, and Ryan~P Adams.
\newblock Practical {Bayesian} optimization of machine learning algorithms.
\newblock In \emph{Advances in Neural Information Processing Systems}, pages
  2951--2959, 2012.

\bibitem[Wolpert and Macready(1997)]{wolpert1997no}
David~H Wolpert and William~G Macready.
\newblock No free lunch theorems for optimization.
\newblock \emph{Evolutionary Computation, IEEE Transactions on}, 1\penalty0
  (1):\penalty0 67--82, 1997.

\bibitem[Zaharia et~al.(2010)Zaharia, Chowdhury, Franklin, Shenker, and
  Stoica]{zaharia2010spark}
Matei Zaharia, Mosharaf Chowdhury, Michael~J Franklin, Scott Shenker, and Ion
  Stoica.
\newblock Spark: cluster computing with working sets.
\newblock In \emph{Proceedings of the 2nd USENIX conference on Hot topics in
  cloud computing}, pages 1--7, 2010.

\end{thebibliography}

\end{document}